\documentclass[10pt,twocolumn,letterpaper]{article}

\usepackage{iccv}
\usepackage{times}
\usepackage{epsfig}
\usepackage{graphicx}
\usepackage{amsmath}
\usepackage{amssymb}

\usepackage{subfigure}
\usepackage{booktabs} 
\usepackage{diagbox}
\usepackage{multirow}
\usepackage{ifthen}
\usepackage{lineno}
\usepackage{ifthen}
\usepackage{cases}
\usepackage{footnote}
\usepackage{lipsum}
\makesavenoteenv{table}

\usepackage[pagebackref=true,breaklinks=true,letterpaper=true,colorlinks,bookmarks=false]{hyperref}

\iccvfinalcopy 

\def\iccvPaperID{****} 

\ificcvfinal\fi

\begin{document}

\title{Adaptive Boundary Proposal Network for Arbitrary Shape Text Detection}


\author{Shi-Xue Zhang$ ^1 $, Xiaobin Zhu$ ^1 $\thanks{Corresponding author.} ,  Chun Yang$ ^1 $, Hongfa Wang$ ^3 $, Xu-Cheng Yin$ ^{1,2,4} $\\
	{\tt\small $ ^1 $School of Computer and Communication Engineering, University of Science and Technology Beijing}\\
	{\tt\small $ ^2 $USTB-EEasyTech Joint Lab of Artificial Intelligence, \tt\small $ ^3 $Tencent Technology (Shenzhen) Co. Ltd}\\
	{\tt\small $ ^4 $Institute of Artificial Intelligence, University of Science and Technology Beijing}\\
	{\tt\small zhangshixue111@163.com, \{zhuxiaobin, chunyang, xuchengyin\}@ustb.edu.cn, hongfawang@tencent.com}
}

\maketitle
\ificcvfinal\thispagestyle{empty}\fi

\begin{abstract}
   Arbitrary shape text detection is a challenging task due to the high complexity and  variety of scene texts. In this work, we propose a novel adaptive boundary proposal network for arbitrary shape text detection,  which can learn to directly produce accurate boundary for arbitrary shape text without any post-processing. Our method mainly consists of a boundary proposal model and an innovative adaptive boundary deformation model. The boundary proposal model constructed by multi-layer dilated convolutions is adopted to produce prior information (including classification map, distance field, and direction field) and coarse boundary proposals. The adaptive boundary deformation model is an encoder-decoder network, in which the encoder mainly consists of a Graph Convolutional Network (GCN) and a Recurrent Neural Network (RNN). It aims to perform boundary deformation in an  iterative way for obtaining  text instance shape guided by prior information from the boundary proposal model. 
   In this way, our method can directly and efficiently generate accurate text boundaries without complex post-processing. Extensive experiments on publicly available datasets demonstrate the state-of-the-art performance of our method. Code is available at the website: \url{https://github.com/GXYM/TextBPN}.
\end{abstract}


\section{Introduction}
Scene text detection has been widely applied in various applications, such as online education, product search, and video scene parsing. Benefiting from the rapid development of deep learning, text detection methods~\cite{CTPN, textboxes, EAST, RRPN} have achieved impressive performance on images in which text instances are regular shape or aspect ratio. Recently, arbitrary shape text detection has attracted ever-increasing interests for it can well adapt to real applications.

\begin{figure}[ht]
	\begin{minipage}[t]{0.99\linewidth}
	\includegraphics[width=1\linewidth]{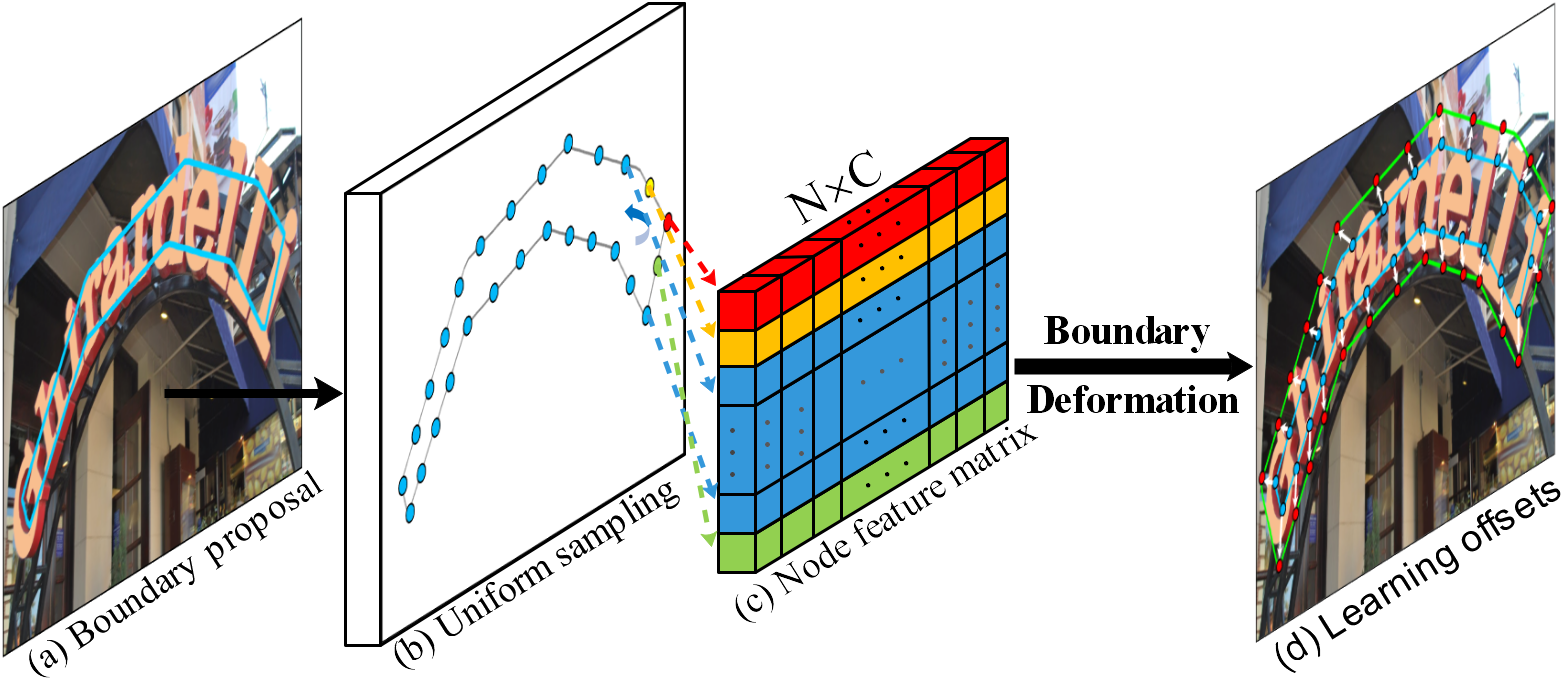}
	\caption{Illustration of the boundary proposal deformation process: (a) Boundary proposal; (b) Sampling on boundaries; (c) Extracting node feature matrix; (d) Learning offsets of sampling vertexes via adaptive boundary deformation model.} \label{fig:fig1}
	\end{minipage}%
\vspace{-1.8em}
\end{figure}

Although arbitrary shape text detection methods~\cite{CRAFT,DB,DRRG} have achieved great improvements in recent years, there are still many issues to be addressed due to the challenging characteristic of scene texts, including varied shape, texture, scale, \etc. Segmentation-based methods~\cite{TextField, CVPR19_PSENet, DB} have sparked a new wave of arbitrary shape text detection that locate text regions by a pixel-level prediction for enhancing the robustness to shape variations. However, there are still two main problems that remain to be explored.

One problem is that segmentation-based methods tend to fail in separating adjacent text instances in image. To solve this problem,  existing methods~\cite{TextSnake, PSENet_v2, CVPR19_LSA,CVPR19_LOMO} usually shrink annotation boundaries as 
kernels (\eg, text kernel~\cite{PSENet_v2}, text center region~\cite{TextSnake}) to distinguish different text instances. For rebuilding a complete text,
these methods usually need to merge the pixels in text regions to kernels by pre-defined expansion rules or auxiliary information (\eg, similarity vector~\cite{PSENet_v2}). 
However, the merging process in~\cite{TextField, CVPR19_PSENet} are always performed by pixel-to-pixel, which is complex and inefficient. The other problem is that the final detected contours of texts in existing segmentation-based methods usually contain a lot of defects and noises. This because the performance of existing segmentation-based methods~\cite{PixelLink, DB, CVPR19_PSENet,PSENet_v2} greatly relies on the accuracy of contour detection, neglecting adaptively adjusting detected contours. Different from generic object instances, text instances usually don't have closed contours and often contain a lot of background
noisy pixels in coarse-grained boundary annotations. These will 
generate unpredictable results of pixels, especially those near boundaries, resulting in a lot of noises and defects in segmentation results.  

To tackle the above-mentioned problems, we propose a novel adaptive boundary proposal network for arbitrary shape text detection, which can learn to directly produce accurate boundary for arbitrary shape text without any post-processing. Our adaptive boundary proposal network is mainly composed of a boundary proposal model and an adaptive boundary deformation model. The boundary proposal model is composed of multi-layer dilated convolutions, which will predict a classification map, a distance field, and a direction field based on shared convolutions. Inspired by RPN~\cite{Faster-rcnn}, we adopt the distance field and pixel classification map to generate coarse boundary proposals as shown in Fig.~\ref{fig:fig1} (a). These coarse boundary proposals can roughly locate texts, and well separate adjacent texts because they are always slimmer than their boundary annotations in our method. To refine the coarse proposals, we adopt an innovative adaptive boundary deformation model to perform iterative boundary deformation for generating accurate text instance shape under the guidance of prior information (classification map, distance field and direction field). For fully excavating and exploiting topology and sequence context in each boundary proposal, the adaptive boundary deformation model adopt an encoder-decoder structure, in which the encoder mainly consists of a GCN and a RNN (B-LSTM). Notably, the proposed method is a unified end-to-end trainable framework with iterative optimization. Extensive experiments demonstrate that our method achieves state-of-the-art performance on several publicly available datasets. 

In summary, the main contributions of this paper are three-fold:
\begin{itemize}
	\item We propose a novel unified end-to-end trainable framework for arbitrary shape text detection, which can directly generate accurate boundaries of arbitrary shape texts without any post-processing.
	
	\vspace{-0.5em}
	
	\item We propose an adaptive boundary deformation model which can perform iterative boundary deformation for refining text boundary.
	
	\vspace{-0.5em}
	
	\item Extensive experiments on public available
	datasets demonstrate the state-of-the-art performance of our method.

\end{itemize}

\begin{figure*}[htbp]
	\begin{center}
	\includegraphics[width=0.99\linewidth]{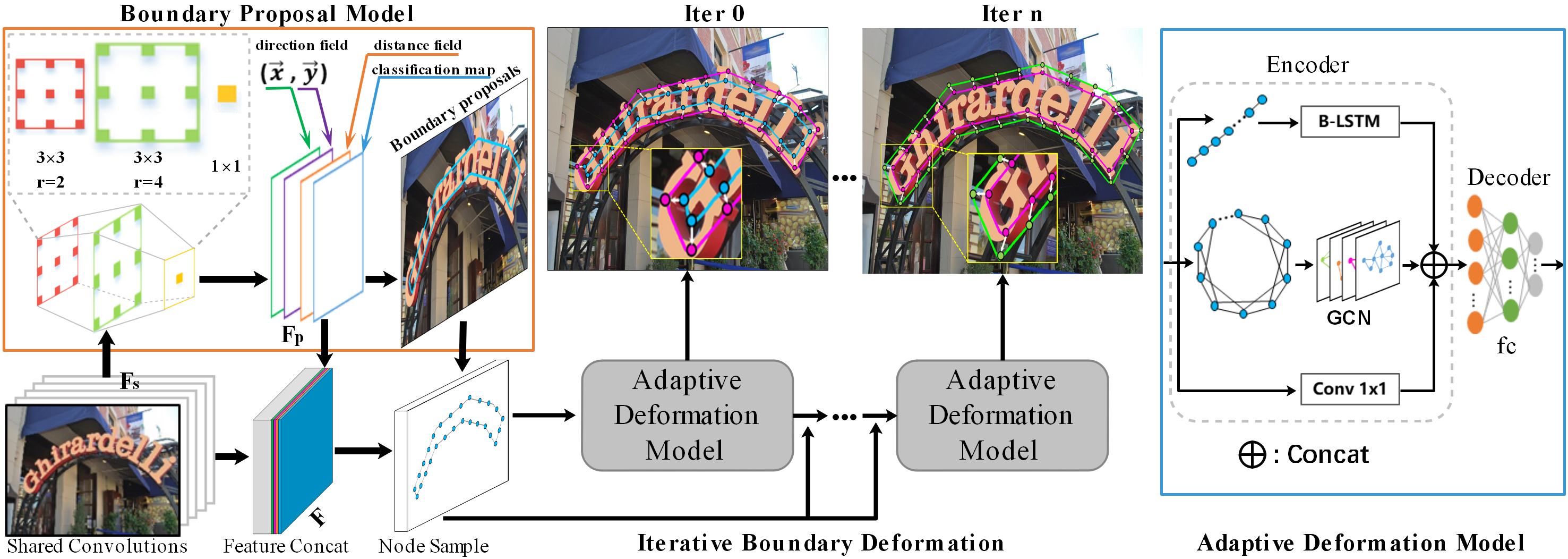}
	\caption{Framework of our method. Our network mainly consists of shared convolutions, 
	boundary proposal model, and adaptive boundary deformation model, which is a unified end-to-end trainable framework with iterative optimization.}
	\label{fig:framework}
	\end{center}
	\vspace{-1.8em}
\end{figure*}

\section{Related Work} \label{Related_Work}


\textbf{Regression-Based Methods.} Methods of this type rely on box-regression based object detection frameworks with word-level and line-level prior knowledge \cite{RRPN,textboxes++,RRD,EAST, CVPR19_LOMO,HAM}. Different from generic objects, texts
are often presented in irregular shapes with various aspect ratios. To deal with this problem, RRPN~\cite{RRPN} and Textboxes++~\cite{textboxes++} localized text boxes by predicting the offsets from anchors. Different from these methods localizing text regions by implementing refinement on pre-defined anchors, EAST~\cite{EAST} and DDR~\cite{DDR} directly regressed the offsets from boundaries or vertexes to the current point for accurate and efficient text detection. Although regression-based methods have achieved good performance in quadrilateral text detection, they often can't well adapt to arbitrary shape text detection.


\textbf{CC-Based Methods.}	The Connected Component (CC) based methods~\cite{Yin-M, SegLink, TextDragon, CRAFT, DRRG} usually detect individual text parts or characters firstly, followed by a link or group post-processing procedure for generating final texts. CRAFT~\cite{CRAFT} detected text regions by exploring affinities between characters. TextDragon \cite{TextDragon} first detected the local region of the text, and then groups them by their geometric relations. Zhang \etal~\cite{DRRG} 
used a graph convolution neural network (GCN) to learn and infer the linkage relationships of text components to group text components. Although CC-based methods have a more flexible representation that can adapt to irregular shape text well, the complex post-processing for grouping text components to complete texts is always time-consuming and unsatisfied.



\textbf{Segmentation-Based Methods.}
Methods of this type \cite{CVPR19_PSENet, TextField, PSENet_v2, DB, MaskTextSpotter} mainly draw inspiration from semantic segmentation methods and detect texts by estimating word bounding areas. To effectively distinguish adjacent text, PSENet~\cite{CVPR19_PSENet} adopted a progressive scale algorithm to gradually expand the pre-defined kernels. In~\cite{PSENet_v2}, Wang \etal proposed an efficient arbitrary shape text detector, named Pixel Aggregation Network (PAN), which is equipped with a low computational-cost segmentation head and learnable post-processing. DB~\cite{DB} performed an adaptive binarization process in a segmentation network, which simplifies the post-processing and enhances the detection performance. However, the performances of these methods are strongly affected by the quality of segmentation accuracy.

	
\textbf{Contour-based methods.} In addition to the above methods, Contour-based methods also have attracted many researchers~\cite{corner, CVPR19_ATRR, Boundary, ContourNet}. Yao \etal~\cite{corner} detected texts by predicting the corner of texts, and  Lyu \etal.~\cite{CVPR19_ATRR} adopted a similar architecture as SSD~\cite{SSD} and rebuilt text with predicted corner points.
Wang \etal.~\cite{Boundary} proposed an end-to-end approach toward arbitrary shape text spotting, which proposed a boundary point detection network to locate the text boundary in the text proposal region. ContourNet~\cite{ContourNet} represented text region with a set of contour points, which adopted a Local Orthogonal Texture-aware Module (LOTM) to model the local texture information of proposal features in two orthogonal directions for generating contour points. However, compared with segmentation-based methods, the contour-based method has a big gap in detection performance and speed without recognition information.

\section{Proposed Method} \label{Proposed_Method}

\subsection{Overview}
\begin{figure}[ht]
	\vspace{-0.5em}	
	\begin{center}
	\includegraphics[width=0.96\linewidth]{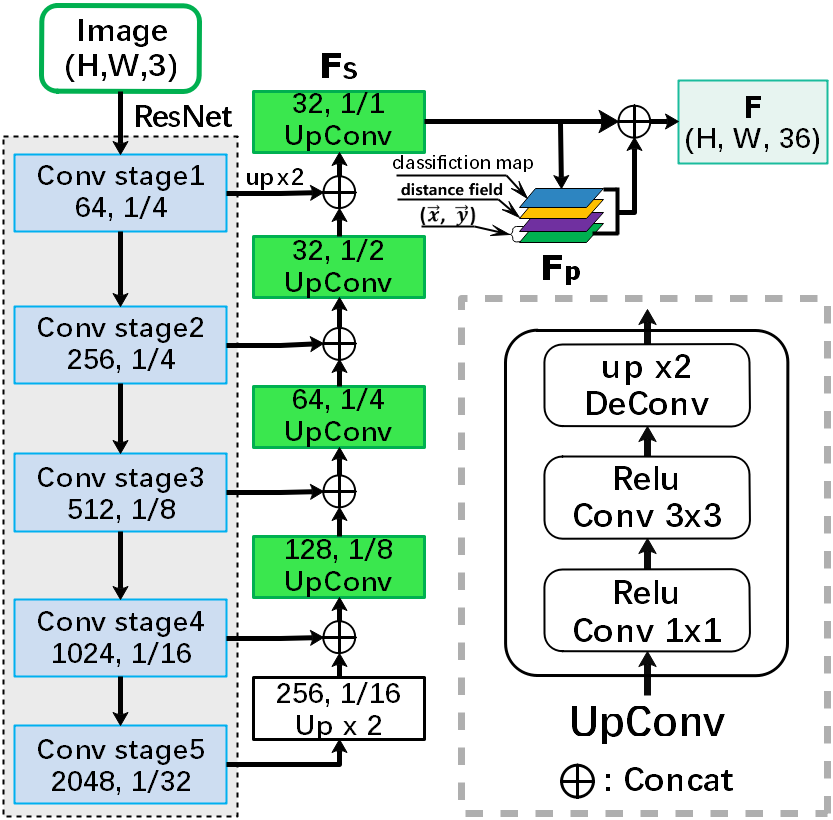}
		\caption{Architecture of shared convolutions, $ F_{S} $ denotes the shared features and $ F_p $ denotes the prior information (classification map, distance field, and direction field).}	
	\label{fig:backbone}
	\end{center}%
	\vspace{-2.0em}
\end{figure}

The framework of our method is illustrated in Fig.~\ref{fig:framework}.
The ResNet-50~\cite{ResNet} is  adopted to extract features. To preserve spatial resolution and take full advantage of multi-level information, we exploit a multi-level
feature fusion strategy (similar to FPN~\cite{FPN}), as shown in Fig.~\ref{fig:backbone}. The boundary proposal model composed of multi-layer dilated convolutions uses the shared features for performing text pixels classification, generating the distance field and direction field~\cite{TextField}. Then, we use these information to produce coarse boundary proposals. Each boundary proposal consists of $ N $ points, representing a possible text instance. For refining the coarse proposals, an adaptive boundary deformation model is proposed to perform iterative boundary deformation for obtaining the more accurate text boundaries under the guidance of prior information (classification map, distance field, and direction field). 

\subsection{Adaptive Boundary Proposal Network}
\begin{figure*}[tbp]
	\begin{center}
		\includegraphics[width=0.98\linewidth]{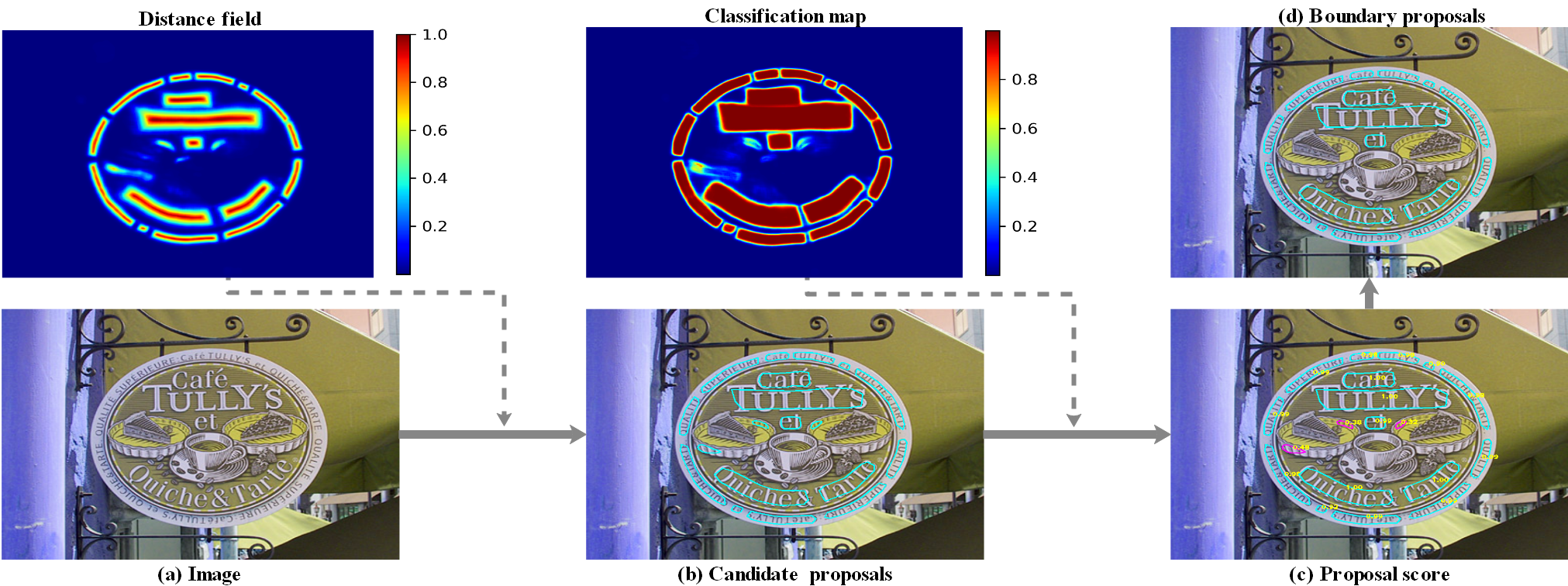}
		\caption{Illustration of boundary proposal generation.}
	\label{fig:proposal}
	\end{center}%
	\vspace{-2.0em}
\end{figure*}

\begin{figure}[tbp]
	\vspace{-0.0em}
	\begin{center}
		\includegraphics[width=0.98\linewidth]{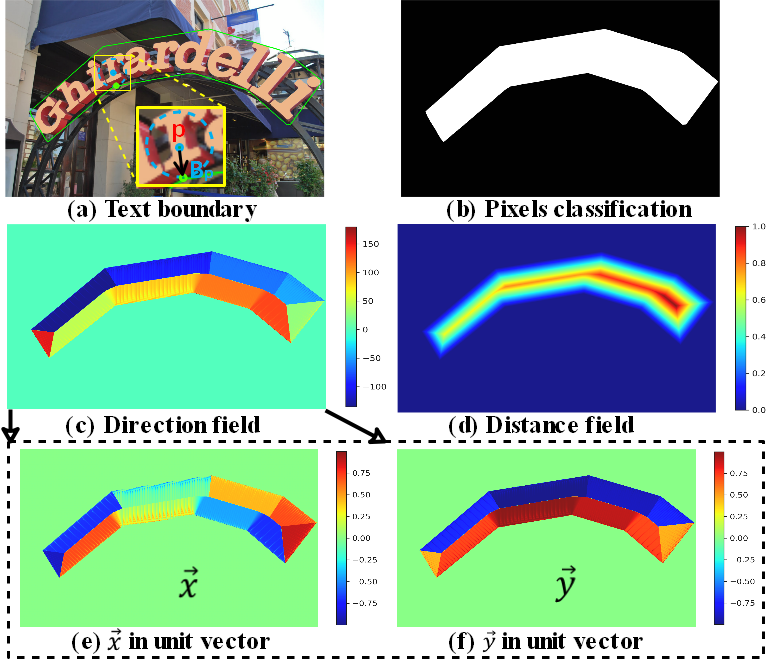}
		\caption{Illustration of ground truths for boundary proposal model, and the unit vector $ (\stackrel{\rightarrow}{x},\stackrel{\rightarrow}{y})$ in (e) and (f) is a vector representation of direction field in (c).}
	\label{fig:field}
	\end{center}%
	\vspace{-2.0em}
\end{figure}

\subsubsection{Boundary Proposal Generation}
The boundary proposal model is composed of multi-layer dilated convolutions, include two $ 3\times 3 $ convolution layers with different dilation rates and one $ 1\times 1 $ convolution layers, as shown in Fig.~\ref{fig:framework}. It will use the shared features extracted from the backbone network to generate classification map, distance field map, and direction field map.

Similar to other text detection methods~\cite{EAST, TextSnake,DRRG}, the classification map contains a classification confidence of each pixel (text/non-text). As in~\cite{TextField, Super_BPD}, the direction field map ($ V $) consists of a two-dimensional unit vector ($ \stackrel{\rightarrow}{x}
,\stackrel{\rightarrow}{y}$), which indicates the direction of each text pixel in boundary to its nearest pixel on boundary (pixel-to-boundary). For each pixel $ p $ inside a text instance $ T $ ,  we will find its nearest pixel $ B_{p} $ on text boundary, as shown in Fig.~\ref{fig:field}. Then, a two-dimensional unit vector $ \mathcal{V}_{gt}(p) $ that points away from
the text pixel $ p $ to $ B_{p} $ can be formulated as 
\\[-2mm]
\begin{small}
\begin{equation}
\mathcal{V}_{gt}(p) \; = \; \left\{\begin{matrix}\ \overrightarrow{B_p p}/\left\vert\overrightarrow{B_p p}\right\vert, & p\in \mathbb{T} 
\\ \\
(0,0), & p \not\in \mathbb{T} \end{matrix}\right.\label{eq:dir}
\end{equation}
\end{small}
where $\left\vert\overrightarrow{B_p p}\right\vert$ represents the distance between $ B_p $ and text pixel $ p $, and $\mathbb{T}$ represent the total set of text instances in an image. For the non-text area ($ p \not\in \mathbb{T} $), we represent those pixels with $(0,0)$.  The  unit vector $ \mathcal{V}_{gt}(p) $ not only directly encodes approximately relative location of $ p $ inside $ T $ and highlights the boundary between adjacent text instances~\cite{TextField}, but also provides direction indication information for boundary deformation.

For boundary deformation, the relative position distance information is as important as the direction information. In this work, 
the distance field map ($ \mathcal{D} $) 
is a normalized distance map, and the normalized distance of the text pixel $ p $ to
nearest pixel $ B_{p} $ on text boundary is defined as
\\[-3mm]
\begin{small}
\begin{equation}
\mathcal{D}_{gt}(p) \; = \; \left\{\begin{matrix}\ \left\vert\overrightarrow{B_p p}\right\vert/L, & p\in \mathbb{T} 
\\ \\
0, & p \not\in \mathbb{T} \end{matrix}\right.
\end{equation}
\end{small}
For the non-text area ($ p \not\in \mathbb{T} $), we represent the distance of those pixels with $0$. $ L $ represents the scale of text instance $ T $ where the pixel $ p $ is located, and is defined as
\\[-3mm]
\begin{small}
	\begin{equation}
	L = max(D_{gt}(p)); \quad p\in T,
	\end{equation}
\end{small}
$ \mathcal{D}_{gt}(p) $ not only directly encodes the relative distance of $ p $ inside $ T $ and further highlights the boundary between adjacent text instances, but also provides a relative distance information for boundary deformation.

With the distance field map ($ \mathcal{D} $), we can generate candidate boundary
proposals by using a fixed threshold ($ th_d $) to the predicted distance, as shown in Fig.~\ref{fig:proposal} (b). However, these candidate boundary proposals inevitably contains false detections. Hence, we calculate the average confidence of each candidate boundary proposal according to the classification map for removing some of them with low confidence ($ th_s $), as shown in Fig.~\ref{fig:proposal} (c) and (d).

\subsubsection{Adaptive Boundary Deformation}
Inspired by interactive annotation of object segmentation methods~\cite{Polygon_RNN++,Curve_GCN} and instance segmentation method~\cite{DeepSnake}, we perform arbitrary shape text detection by
deforming a boundary proposal to a more accurate text boundary.
Specifically, we take a boundary as input based on CNN features and predict per-vertex offsets pointing to the text boundary. In~\cite{Polygon_RNN++}, the authors 
use a Recurrent Neural Network to model the sequence of 2D vertices of the polygon outlining an object.
Afterwards, they propose a method~\cite{Curve_GCN} to treat the location of each control point as a continuous random variable,
and learn to predict these variables via a Graph Neural Network that takes image evidence as input. Inspired by ~\cite{Polygon_RNN++,Curve_GCN}, DeepSnake~\cite{DeepSnake} performs object segmentation by
deforming an initial contour to object boundary with circular convolution which consists of multi-layer 1-D convolutions with $ 1 \times N $ kernel size.
However, these methods only consider individual topology context or sequence context, which isn't highly  satisfactory for text detection. For the unique properties of texts, both topology context and sequence
context are important in detection. 

For each boundary proposal, we will uniformly sample $ N  $ control points for facilitating batch processing. As shown in~Fig.~\ref{fig:framework}, the sampled control points form a closed polygon in which topology context and sequence context are co-existing. To fully exploit the boundary topology and sequence context, we introduce an adaptive boundary deformation model combined with GCN and RNN, which can efficiently perform feature learning and iterative boundary deformation for refining coarse text boundaries.

Let $ cp_i = [x_i, y_i]^{T} $ denote the location of the $ i$-th control point, and $ p = \{{p}_0,{p}_1,...,{p}_{N-1}\} $ be the set of all control points. For a boundary proposal with $ N  $ control points, we first construct feature
vectors for each control point. The input feature $ f_{i} $ for a control point $ cp_i $ is a concatenation of  32-D shared features $ F_s $ obtained by CNN backbone and 4-D prior features $ F_p $ (\eg, pixels classification, distance field and direction field). Therefore, the features of a control point are extracted from the corresponding location in $ F:f_{i} = concat\{F_s(x_i,y_i),F_p(x_i,y_i)\}$. Here, $ F_s(x_i,y_i) $ and  $ F_p(x_i,y_i) $ are computed by bilinear interpolation.

After obtaining the feature matrix $ X $ (size: $ N \times C $)  of boundary proposal, we adopt the adaptive deformation model based on an encoder-decoder architecture to perform efficient feature learning and iterative boundary deformation. The encoder model is combined with GCN and RNN for feature learning, which can fully exploit and fuse the boundary topology and sequence context. As shown in Fig.~\ref{fig:framework}, the encoder model can be formulated as
\\[-3mm]
\begin{small}
	\begin{equation}
	X^{'} = RNN(X) \oplus GCN(X) \oplus Conv_{1 \times 1}(X)
	\end{equation}
\end{small}
where ``$ \oplus $'' refers to the concatenation operation; RNN is consisted of one-layer B-LTSM with 128 hidden size; $ Conv_{1 \times 1} $ consists of  one-layer $ 1 \times 1 $ convolution layers with 128 dimensions, which form a residual connection like RestNet~\cite{ResNet}; GCN is consisted of four graph convolution layers activated by ReLU, and the graph convolution layer in our method is formulated as
\\[-3mm]
\begin{small}
	\begin{gather}
	\bold X_g = ReLU((\bold X \oplus \bold G\bold X)\bold W),\\
	\bold G = \bold{\tilde{D}}^{-1/2}\bold{\tilde{A}}\bold{\tilde{D}}^{-1/2},
	\end{gather}
\end{small}
where $ \bold {X} \in \Re^{N \times d_{i}} , \bold X_g \in \Re^{N \times d_{o}}  $, $ d_i/d_o $ is the dimension of in/out features, and $ N  $ is the number of control points; $ \bold G $ is a symmetric normalized Laplacian of
size $ N \times N $; $ W $ is a layer-specific trainable weight matrix; $ \tilde{A} = A + I_N $ is an adjacency matrix of the local graph with added self-connections; $ I_N $ is the identity matrix and $ \bold{\tilde{D}} $ is a diagonal matrix with $ \tilde{D}_{ii} = \sum_{j}\tilde{A}_{ij}$. We form
$ A $ by connecting each control point in $ p $ with its four neighbors. 

The decoder in adaptive deformation model consists of three-layer $ 1 \times 1 $ convolutions with ReLU, which will learn to predict offsets between control points and the target points. To obtain the more accurate text boundary, we perform iterative boundary deformation, as shown in Fig.~\ref{fig:framework}.

\subsection{Optimization}
\noindent In this work, the total loss $ \mathcal{L} $ can be formulated as \\[-3mm]
\begin{small}
\begin{equation}
\mathcal{L} = \mathcal{L}_{Bp} + \dfrac{\lambda * \mathcal{L}_{Bd}}{1+e^{(i-eps)/eps}},\label{tal_loss}
\end{equation}
\end{small}
where $ \mathcal{L}_{Bp} $ is a loss for the boundary proposal model,  and $ \mathcal{L}_{Bd} $ is a loss for the adaptive boundary deformation model; $ eps $ denotes the maximum epoch of training, and $ i $ denote the $ i$-th epoch in train. In our experiments, $ \lambda  $ is set to 0.1. In Eq.~\ref{tal_loss}, $ \mathcal{L}_{Bp} $ is computed as \\[-3mm]
\begin{small}
\begin{equation}
\mathcal{L}_{Bp} = \mathcal{L}_{cls} + \alpha* \mathcal{L_{D}}+ \mathcal{L_{V}},
\end{equation}
\end{small}
where $ \mathcal{L}_{cls} $ is a cross-entropy classification loss for pixels classification, and $ \mathcal{L_{D}} $ is a $ L_{2} $ regression loss for distance field. OHEM~\cite{OHEM} is adopted for $ \mathcal{L}_{cls} $ and $ \mathcal{L_{D}} $ in which the ratio between the negatives and positives is set to 3:1.  To balance the losses in $ \mathcal{L}_{Bp} $, the weights $ \alpha $ is  set to 3.0. 
Similar to~\cite{Super_BPD}, $ \mathcal{L_{V}} $  consists of $ L_2 $-norm distance and angle distance for direction field $ \mathcal{V} $:\\[-3mm]
\begin{small}
\begin{equation}
\mathcal{L_{V}} = \sum_{p \in \Omega} {  w(p){\|\mathcal{V}_p - \hat{\mathcal{V}}_p\|}_2 } + \frac{1}{\mathbb{T}}\sum_{p \in \mathbb{T}} (1 - \cos( \mathcal{V}_p, \hat{\mathcal{V}}_p)) ,
\label{eq:finalloss}
\end{equation}
\end{small}
where $ \Omega $ represents image domain; the weight ($w(p) = 1/\sqrt{|GT_p|}$) at pixel $p$ is proportional to the inverse square root of the size of ground truth segment $GT_p$ containing $p$.

$ \mathcal{L}_{Bd} $ is a point matching loss similar to~\cite{Curve_GCN}. In this work, the prediction and ground truth control point sets have equal sized and similar order (counter-clockwise), denoted as $\textbf{p}=\{p_0,p_1,\cdots,p_{N-1}\}$, and $\textbf{p}'=\{p'_0,p'_1,\cdots,p'_{N-1}\}$ ($N$ is the number of points), respectively. Hence, the matching loss for $ \textbf{p} $ and $ \textbf{p}' $
is defined as\\[-3mm]
\begin{small}
\begin{equation}
\mathcal{L}_{(\textbf{p}, \textbf{p}')} = \min_{j \in [0\cdots, N-1]}\sum_{i=0}^{N-1}{smooth_{L1}(p_i, p'_{(j+i)\%N})},
\end{equation}
\end{small}
Because there is usually more than one text instance in an image, $ \mathcal{L}_{Bd} $ is defined as\\[-3mm]
\begin{small}
\begin{equation}
\mathcal{L}_{Bd} = \frac{1}{\mathbb{T}}\sum_{p \in \mathbb{T}} \mathcal{L}_{(\textbf{p}, \textbf{p}')},
\end{equation}
\end{small}
where $\mathbb{T}$ represents all the text instances in an image, $ \textbf{p} $ represents the control point set for text instance $ T $ ($ T \in  \mathbb{T} $).

\section{Experiments} \label{Experiments}
\subsection{Datasets}
\noindent\textbf{Total-Text}: It consists of $1,255$ training and $300$ testing complex images, including horizontal, multi-oriented, and curved text with polygon and word-level annotations.

\noindent\textbf{CTW-1500}: It consists of $1,000$ training and $500$ testing images, and curved text instances are annotated by polygons with 14 vertices.

\noindent\textbf{MSRA-TD500}: It consists of $500$ training and $200$ testing images, including English and Chinese texts which contain multi-lingual long texts with multi-orientations.

\noindent\textbf{SynthText}: It contains 800k synthetic images generated by blending natural images with artificial text which are all word-level annotated .

\noindent\textbf{ICDAR2017-MLT}: It consists of $7,200$ training images, $1,800$ validation images, and $9,000$ test images with multi-lingual ($9$ languages) texts annotated by quadrangle.

\newif\ifnoFPS
\noFPStrue
\ifnoFPS

\begin{table*}[htbp]
	\begin{center}
		\renewcommand{\arraystretch}{1.0}
		\caption{Ablation experiments for deformation model on Total-Text and CTW-1500. The best score is highlighted in \textbf{bold}.}	\label{table:Ablation}
		\begin{tabular}{||c||c|c|c|c||c|c|c|c||}
			\hline
			\multicolumn{1}{||c||}{ \multirow{2}*{ \textbf{Methods}}}
			& \multicolumn{4}{c||}{\textbf{   Total-Text}} 
			& \multicolumn{4}{c||}{\textbf{ CTW-1500}}\\
			\cline{2-9}
			&\textbf{Recall}
			& \textbf{Precision} & \textbf{F-measure}&
			\textbf{FPS}&
			\textbf{Recall}& 
			\textbf{Precision} & \textbf{F-measure}&
			\textbf{FPS}\\
			\hline
			FC &81.56 &90.16 &85.65&9.52&78.32 &85.03 &81.54&11.13\\
			
			RNN &\textbf{83.31} &87.71 &85.93&11.15&\textbf{81.26} &86.00 &83.56&12.22\\
			Circular convolution &82.80 &89.73 &86.13 &9.33&80.35 &84.88 &82.55&10.89 \\
			Graph convolution &82.74 &89.94 &86.19  &10.42 &80.31&86.12 &83.12&11.94\\
			
			
			\textbf{Adaptive deformation}&83.30 &\textbf{90.76} &\textbf{86.87} &10.56&80.57 &\textbf{87.66} &\textbf{83.97}&12.08 \\
			\hline
		\end{tabular}
	\end{center}%
\vspace{-1.8em}
\end{table*}

\subsection{Implementation Details}\label{trainstep}
In our experiments, we first pre-train our network on SynthText by one epochs, in which images are randomly cropped and resized to $ 512 \times 512 $. In pre-training, the Adam~\cite{ADAM} optimizer is applied  with a fixed learning rate of $ 0.001 $, and a mini-batch is set to 16. In fine-tuning, we randomly crop the text region, and resize them to $ 640 \times 640 $ for training the model with the mini-batch 12. The Adam~\cite{ADAM} optimizer is adopted, in which the initial learning rate is $ 0.001 $ and decreased to $ 0.9 $ of the original after each 50 epochs. The data augmentation includes: random rotation with an angle (sampled by Gaussian distribution in ($-60^{\circ}, 60^{\circ}$)), random cropping, and random flipping. In inference, we keep the aspect ratio of test images, then resize and pad them into the same size for testing.  The code is implemented with PyTorch 1.7 and python 3. Training is performed on single GPU (RTX-3090), and testing is performed on single GPU (GeForce RTX-2080) with Intel Xeon Silver 4108 CPU @ 1.80GHz.

\subsection{Ablation Study}\label{exp_ablation_aam}
In ablation experiments, we only train the model on  corresponding real-world datasets for 660 epochs without pre-training, and the other training settings are identical with  the fine-tuning process in Sec~\ref{trainstep}. In testing, the short side of an image is scaled to 640, and ensure that the long side doesn't exceed 1,024. The threshold $ th_d $ and $ th_s $ are set to 0.3 and 0.8, respectively.

\noindent\textbf{Effectiveness of adaptive deformation model.}
To verify the effectiveness of the adaptive deformation model, we conduct ablation experiments on Total-Text and CTW-1500. Our deformation model consists of an encoder and a decoder. For fair comparison, we 
use a lightweight full connection network (FC) structure as decoder, and we adopt four types of encoder, \ie, FC with $ Conv_{1 \times 1} $, RNN, circular convolution, and graph convolution (GCN), for conducting comparative experiments. As listed in Tab.$ \, $\ref{table:Ablation}, our adaptive deformation model achieves the best performance compared with the other four methods on both Total-Text and CTW-1500, which achieves improvements by $0.94 \%$ in terms of F-measure on Total-Text compared with RNN, and  by $0.85 \%$ in terms of F-measure on Total-Text compared with GCN. Moreover, our adaptive deformation model doesn't bring obvious more consume of detection time.

\begin{figure}[htbp]
	\begin{center}
		\includegraphics[width=0.98\linewidth]{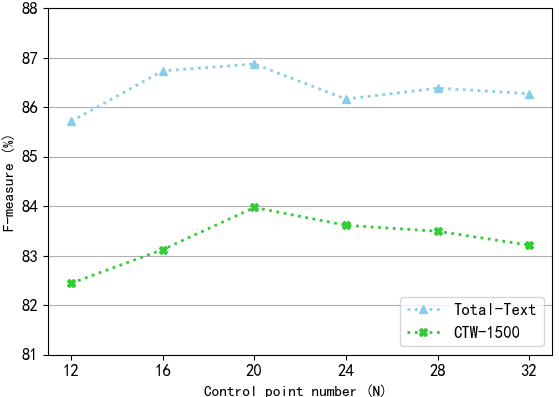}
		\caption{Experimental results of control point number (N).}
		\label{fig:ctr_pts}
	\end{center}
\vspace{-2.0em}
\end{figure}

\begin{figure*}[htbp]
	\subfigcapskip=2pt
	\centering
	\subfigure[boundary proposals]{
		\begin{minipage}[t]{0.245\linewidth}
			\centering
			\includegraphics[width=4.3cm,height=3cm]{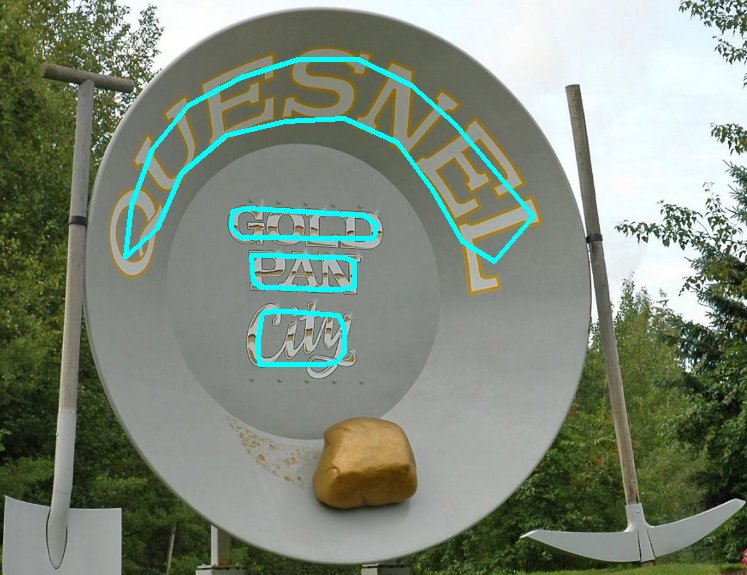}\\
			\includegraphics[width=4.3cm,height=3cm]{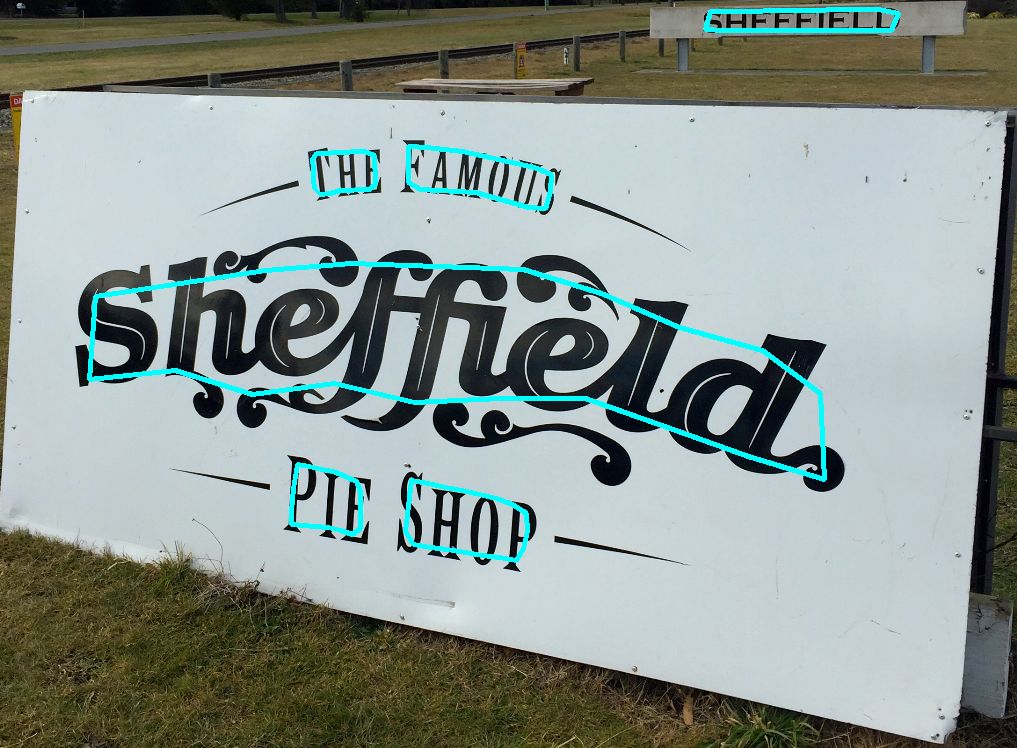}
		\end{minipage}%
	}%
	\subfigure[iter 1]{
		\begin{minipage}[t]{0.245\linewidth}
		\centering
		\includegraphics[width=4.3cm,height=3cm]{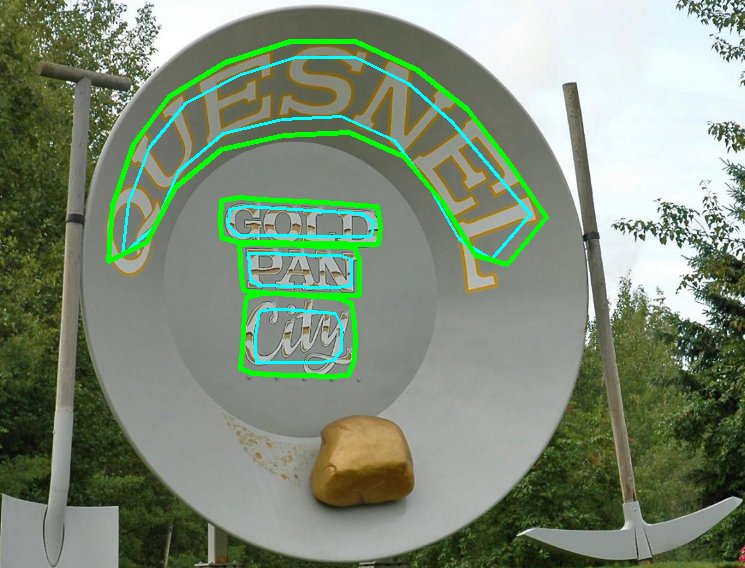}\\
		\includegraphics[width=4.3cm,height=3cm]{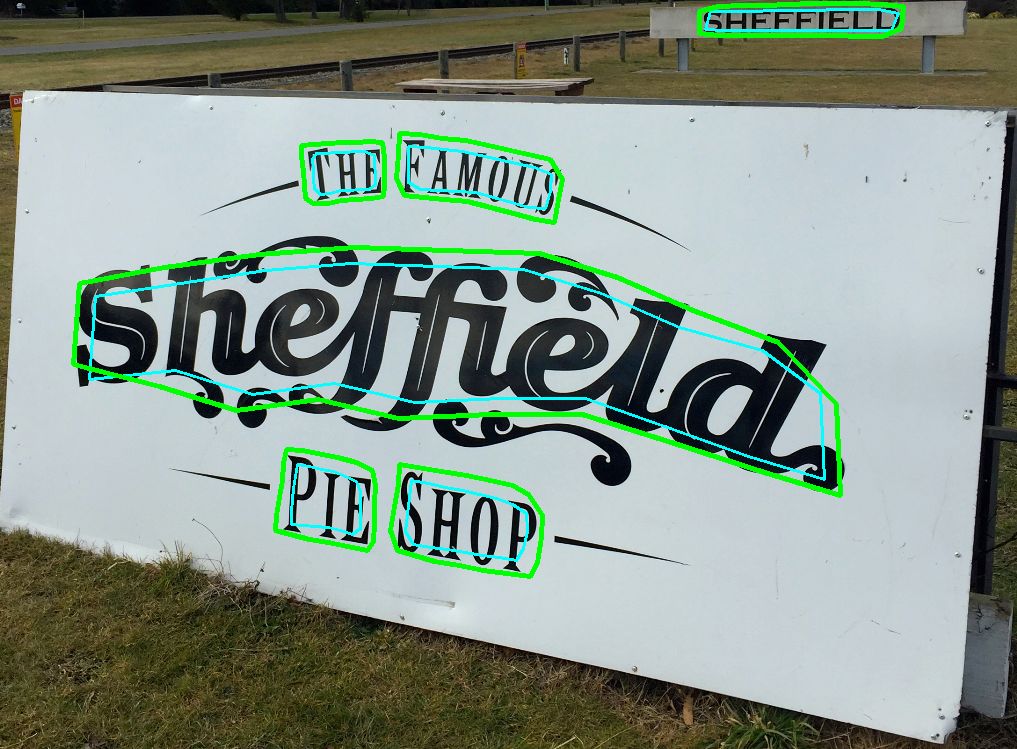}
		\end{minipage}%
	}%
	\subfigure[iter 2]{
		\begin{minipage}[t]{0.245\linewidth}
			\centering
			\includegraphics[width=4.3cm,height=3cm]{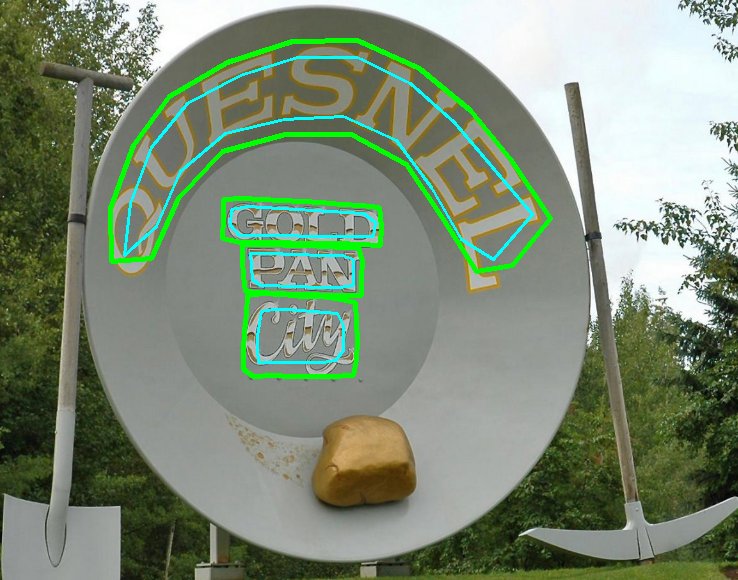}\\
			\includegraphics[width=4.3cm,height=3cm]{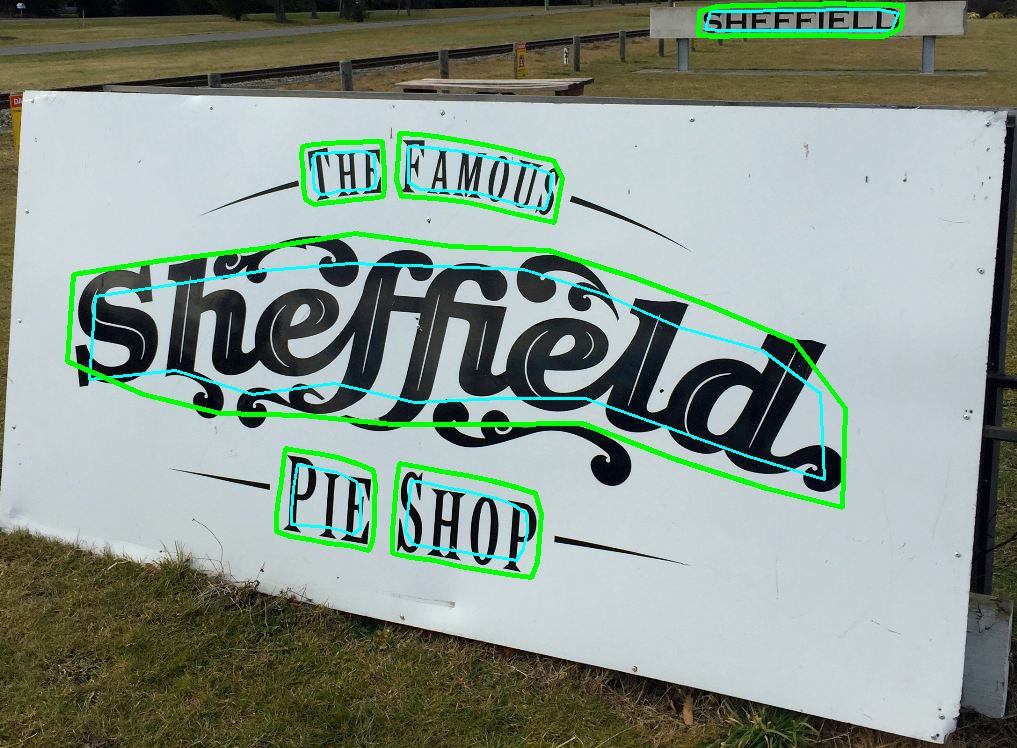}
		\end{minipage}%
	}%
	\subfigure[iter 3]{
		\begin{minipage}[t]{0.245\linewidth}
			\centering
			\includegraphics[width=4.3cm,height=3cm]{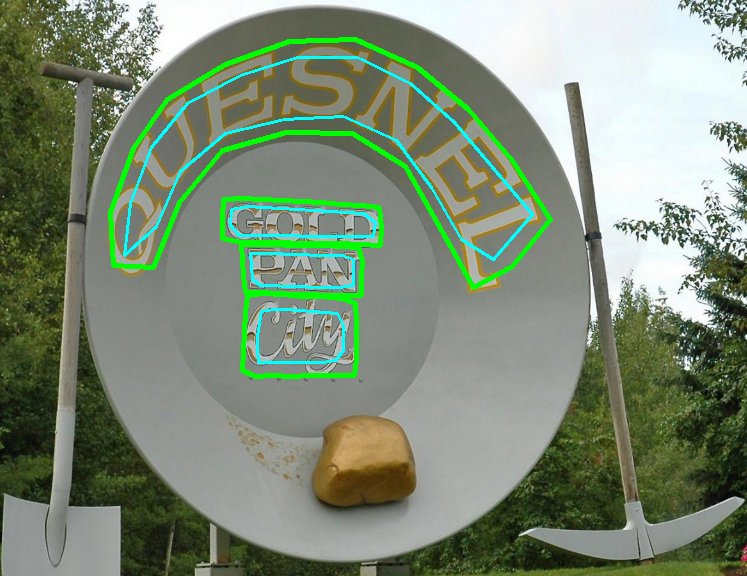}\\
			\includegraphics[width=4.3cm,height=3cm]{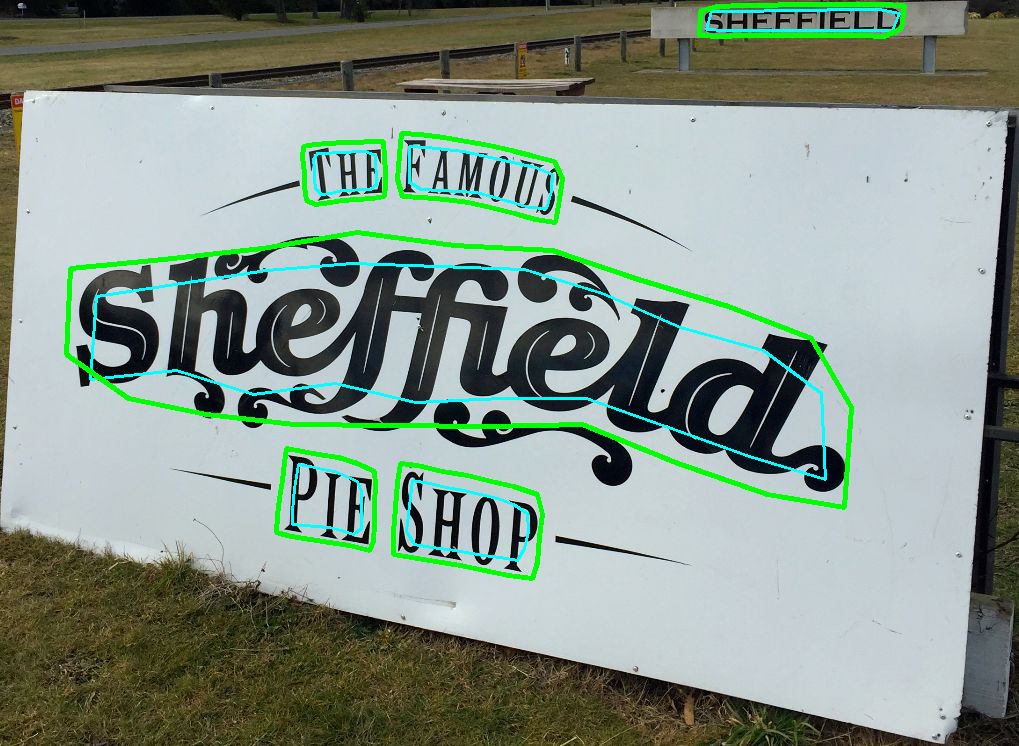}
		\end{minipage}%
	}%
	\centering
	\caption{ Visual results of different iterations. The blue contours are boundary proposals, and the green contours are detection boundaries after iterative deformation.}
	\label{fig:iter_fig}
	\vspace{-1.2em}
\end{figure*}
\medskip
\noindent\textbf{Influence of control point number ($ N $).}
We study the influence of the control point number ($ N $) by setting the number of control point number from 12 to 32 with an interval of 4. The models are evaluated on Total-Text and CTW-1500. From Fig.~\ref{fig:ctr_pts}, we can find that the F-measures drops when $ N $ is too large or too small. Especially, too few control points will make great degradation of performance. This is because the detection boundary often can’t correctly cover the whole text when the control number is too small. Apparently, our model achieves the best detection performance on both two datasets  while the number of control points is around 20. Therefore, in our experiments, the number of control points is fixed at 20.

\begin{table}[ht]
\renewcommand{\arraystretch}{0.9}
\vspace{-0.5em}
\caption{Experimental results of different iterations on CTW-1500.}
\begin{tabular}{r|p{48 pt}<{\centering}|p{48 pt}<{\centering}|p{ 48 pt}<{\centering}}
& Iter. 1 & Iter. 2 & Iter. 3\\
\midrule[1.5pt]
F-measure & 82.24 & 83.33 & \textbf{83.97} \\
Speed (fps) & \textbf{13.68 }& 12.83 & 12.08\\
\end{tabular}
\vspace{-1.0em}
\label{table:iterative}
\end{table}

\noindent\textbf{Influence of iteration number.}
To fully validate the influence of iteration number, we
further compare models with different inference iterations. As listed in Tab.~\ref{table:iterative}, with the increase of the number of iterations, the detection performance is gradually improved but the inference speed of our model is gradually dropped. When the number of iterations is from 2 to 3, the increase of detection performance is not very obvious. Considering the balance of speed and performance, the number of iterations is set to 3 by default in our experiments. As shown in Fig.~\ref{fig:iter_fig}, the detection boundaries become more accurate along with the increase of iterations.

\newif\ifnoFPS
\noFPStrue
\ifnoFPS
\begin{table}[htbp]
	\begin{center}
	\renewcommand{\arraystretch}{0.95}
	\caption{Ablation study for classification map (cls), distance field (dis), and direction field (dir) on Total-Text.} 
	\label{table:Ablation_field}
	\begin{tabular}{ccc|ccc}
		\hline
		{\textbf{cls}}&{\textbf{dis}} & {\textbf{dir}} & \textbf{Recall}& \textbf{Precision} & \textbf{F-measure} \\
		\hline
		{$ \checkmark $}&{$ \times $} &{$ \times $}  &76.96 &83.01  &79.87\\
		{$ \checkmark $} &{$ \checkmark $}&{$ \times $} &81.97 &88.95 &85.32\\
		{$ \checkmark $}&{$ \checkmark $}  &{$ \checkmark $} &\textbf{83.30} &\textbf{90.76} &\textbf{86.87} \\
		\hline
	\end{tabular}
    \end{center}%
\vspace{-2.0em}
\end{table}
\noindent\textbf{Influence of prior information.}
We conduct ablation studies on Total-Text to verify the
importance of each prior information (\eg, classification map, distance field and direction field). As listed in Tab.~\ref{table:Ablation_field}, the detection performance is unsatisfactory when only use classification map. Once distance field and direction field are introduced, the performance is improved significantly, and F-measure is improved by $5.45\%$ and $1.55\%$, respectively.

\begin{small}
	\begin{table}[htbp]
		\begin{center}
			\renewcommand{\arraystretch}{0.95}
			\caption{Experimental results on Total-Text for different resolution FPN. “R”, “P”, and “F” represent Recall, Precision, and F-measure, respectively.}	\label{table:fpn}
			\begin{tabular}{c|cccc}
				\hline
				\textbf{Methods}
				&\textbf{R}
				& \textbf{P} & \textbf{F}&
				\textbf{FPS}\\
				\hline
				FPN-P1 ($ {1}/{1} $)&\textbf{83.30} &\textbf{90.76} &\textbf{86.87} &10.56\\
				FPN-P1 ($ {1}/{2} $)&82.63 &90.75 &86.50 &12.68\\
				FPN-P2 ($ {1}/{4} $)&82.99 &89.51 &86.13 &\textbf{15.17}\\
				\hline
			\end{tabular}
		\end{center}%
		\vspace{-1.5em}
	\end{table}
\end{small}
\noindent\textbf{Influence of different resolution FPN.} We have conducted experiments without any pre-training to explore the influence of 
using the different resolution FPN-layer
as the shared features. In Tab.~\ref{table:fpn}, FPN-P2 ($ {1}/{4} $) means that we use the FPN-P2 layer
as the shared features (the resolution of FPN-P2 is $ {1}/{4} $ of the original document image). From Tab.~\ref{table:fpn}, we can see that FPN-P1 and FPN-P2 both achieve the state-of-the-art performance on Total-Text.

\begin{figure*}[htbp]
\subfigcapskip=2pt
\centering
\subfigure[Total-Text]{
	\begin{minipage}[t]{0.245\linewidth}
		\centering
		\includegraphics[width=4.3cm,height=3cm]{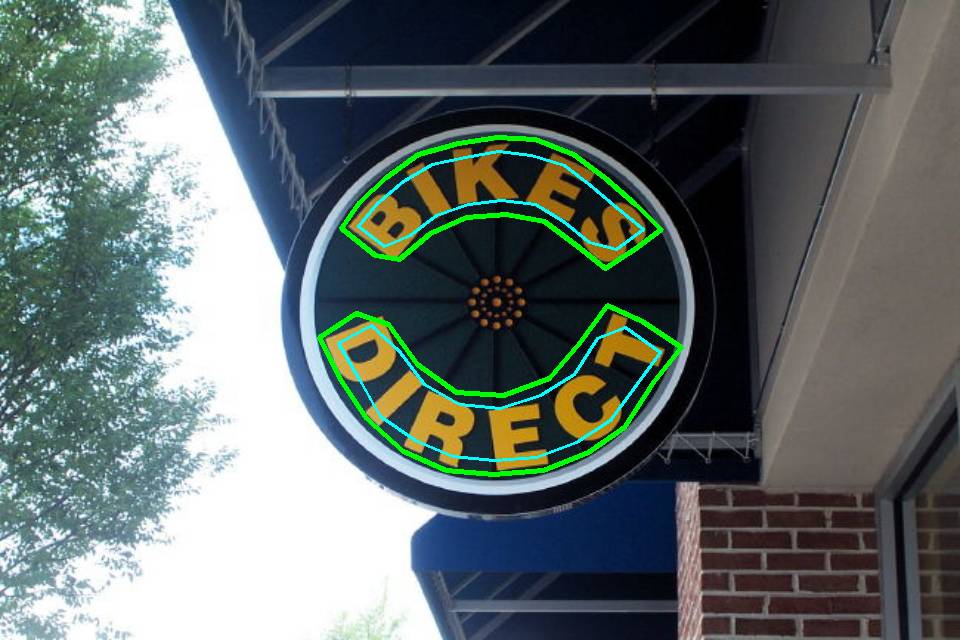}\\
		\includegraphics[width=4.3cm,height=3cm]{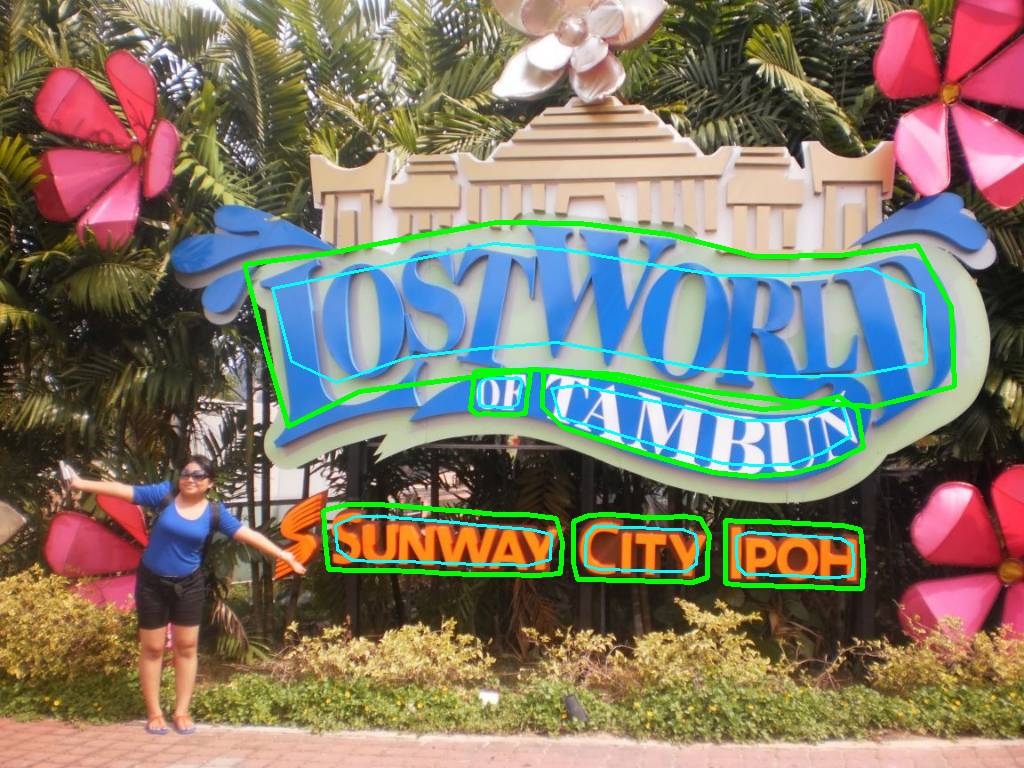}
	\end{minipage}%
}%
\subfigure[Total-Text]{
	\begin{minipage}[t]{0.245\linewidth}
		\centering
		\includegraphics[width=4.3cm,height=3cm]{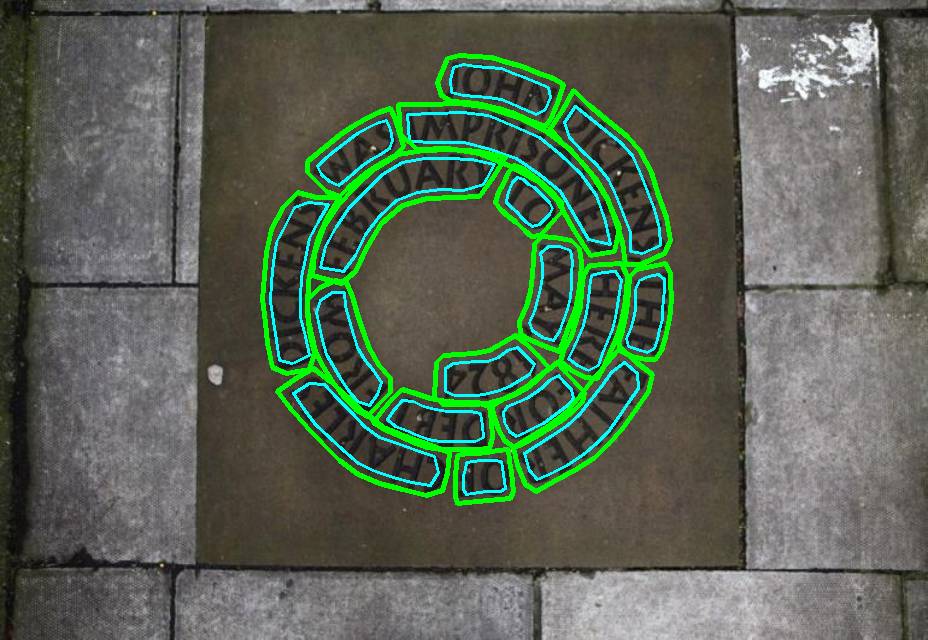}\\
		\includegraphics[width=4.3cm,height=3cm]{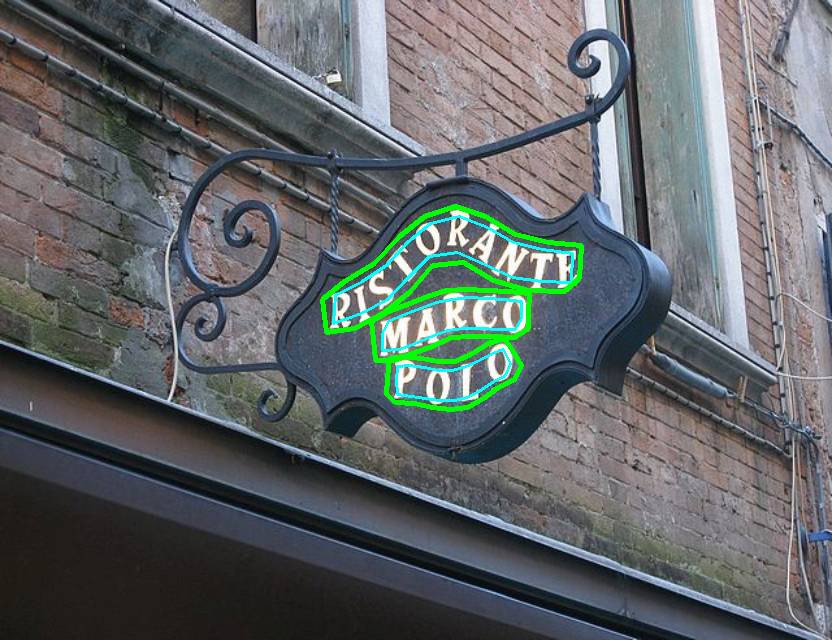}
	\end{minipage}%
}%
\subfigure[CTW-1500]{
	\begin{minipage}[t]{0.245\linewidth}
		\centering
		\includegraphics[width=4.3cm,height=3cm]{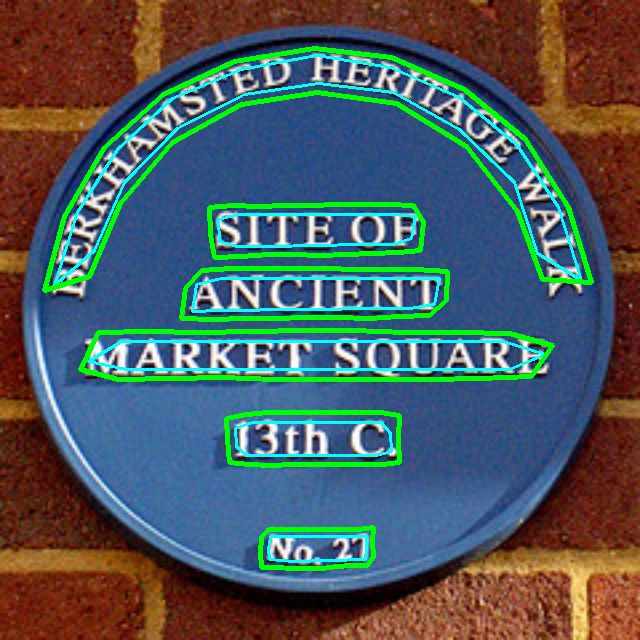}
		\includegraphics[width=4.3cm,height=3cm]{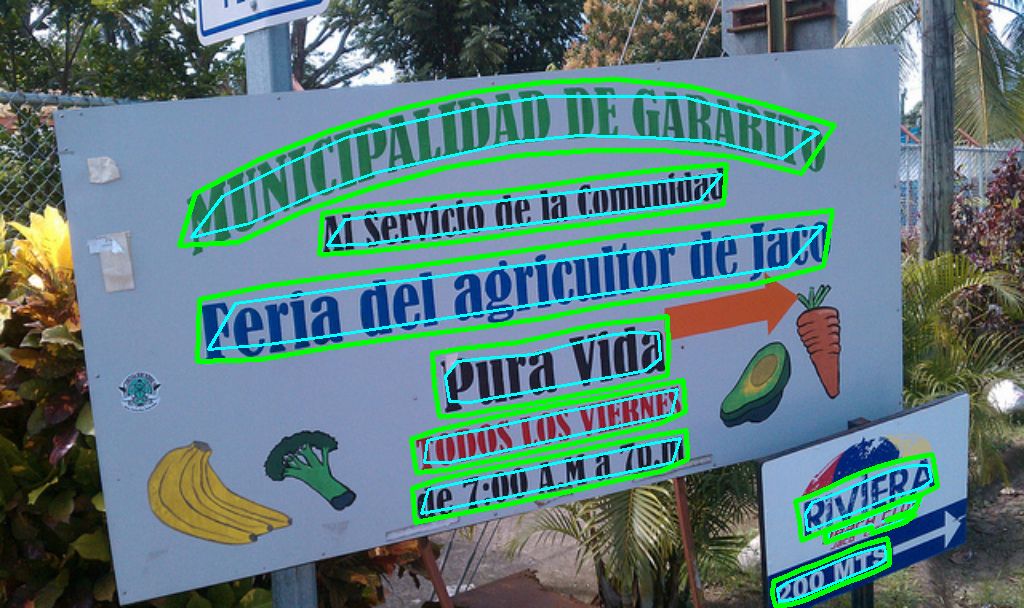}
	\end{minipage}%
}%
\subfigure[CTW-1500]{
	\begin{minipage}[t]{0.245\linewidth}
		\centering
		\includegraphics[width=4.3cm,height=3cm]{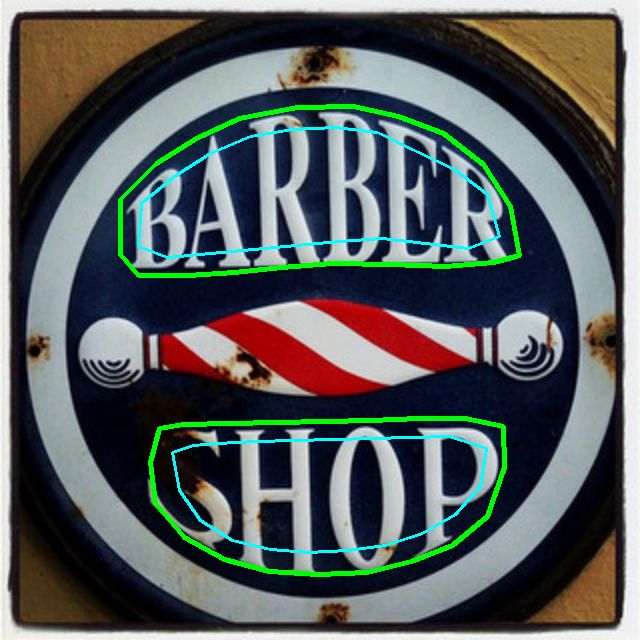}
		\includegraphics[width=4.3cm,height=3cm]{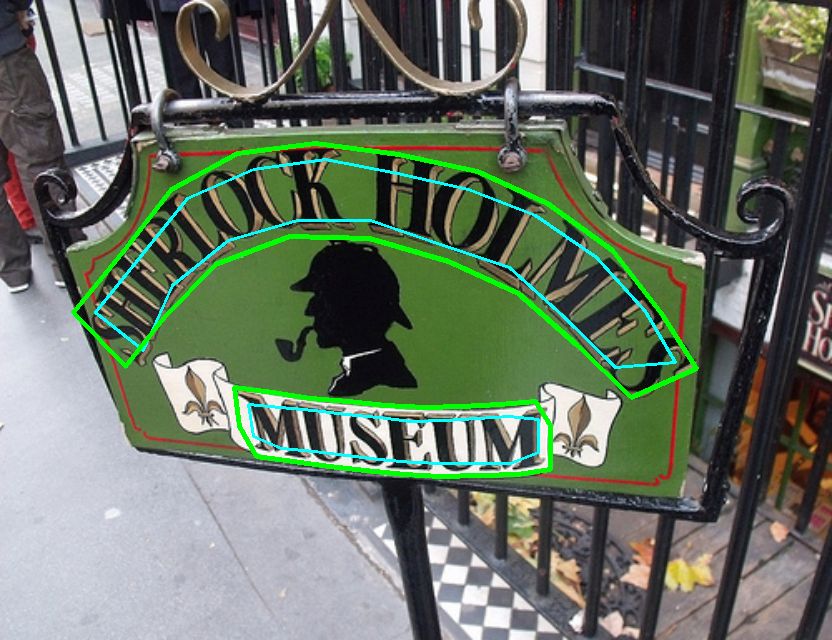}
	\end{minipage}%
}%
	
\centering
\caption{ Visual experimental results. The blue contours are boundary proposals, and the green contours are final detection boundaries.}
\label{fig:result_vis}
\vspace{-1.2em}
\end{figure*}
\begin{table}[htbp]
	\begin{center}
	\renewcommand{\arraystretch}{0.9}
	\caption{Experimental results on Total-Text. `Ext' denotes that the extra pre-training data is used. `Syn'/ `MLT' means SynthText or ICDAR2017-MLT dataset are used for pre-training, and `MLT+' means there are other pre-training data besides MLT.}
	\label{table:TotalText}
	\begin{tabular}{m{2.5cm}<{\centering}
			|m{0.75cm}<{\centering}
			|m{0.6cm}<{\centering}
			m{0.6cm}<{\centering}
			m{0.6cm}<{\centering}
			m{0.5cm}<{\centering}}
		\toprule
		Methods&Ext&R& P& F& FPS \\
		\hline
		TextSnake \cite{TextSnake} &Syn&74.5 &82.7 & 78.4&-\\
		ATTR \cite{CVPR19_ATRR} &-&76.2&80.9 & 78.5&10.0\\
		MSR \cite{MSR} &Syn &85.2  &73.0 & 78.6&4.3\\
		CSE \cite{CVPR19_CSE} &MLT&79.7 &81.4 & 80.2&0.42\\
		TextDragon \cite{TextDragon} &MLT+&75.7 &85.6 & 80.3&-\\
		TextField \cite{TextField}  &Syn&79.9 &81.2 & 80.6&6.0\\
		PSENet-1s \cite{CVPR19_PSENet} &MLT&77.96 &84.02 & 80.87&3.9\\
		SegLink++ \cite{SegLink++} &Syn&80.9 &82.1 & 81.5&-\\
		LOMO \cite{CVPR19_LOMO} &Syn&79.3 &87.6 & 83.3&4.4\\
		CRAFT \cite{CRAFT} &Syn&79.9 &87.6 & 83.6&-\\
		DB \cite{DB}&Syn&82.5 &87.1  &84.7&32.0\\
		PAN \cite{PSENet_v2} &Syn&81.0 &89.3  &85.0&\textbf{39.6}\\
		TextPerception\cite{TextPerception} &Syn&81.8&88.8&85.2&-\\
		ContourNet~\cite{ContourNet}&-&83.9&86.9&85.4&3.8\\
		DRRG~\cite{DRRG}&MLT&84.93&86.54&85.73&-\\
		Boundary~\cite{Boundary}&Syn&85.0 &88.9 &87.0&-\\
		\hline
		\textbf{Ours}&-&83.30&\textbf{90.76} &86.87 &10.56\\
		\textbf{Ours}&Syn&84.65 &90.27 & 87.37&10.28\\
		\textbf{Ours} & MLT&\textbf{85.19} & 90.67 & \textbf{87.85}&10.69\\
		\bottomrule
	\end{tabular}
    \end{center}%
\vspace{-2.0em}
\end{table}

\subsection{Comparison with the state-of-the-arts}
We compare our method with the previous state-of-the-art methods on Total-Text, CTW-1500 and MSRA-TD500. In testing, the short side of the images is scaled to 640, and the long side will be suppressed in 1,024. The threshold $ th_d $ for 
obtaining the candidate boundary proposals with distance filed is fixed to 0.3.

\begin{table}[htbp]
	\begin{center}
	\renewcommand{\arraystretch}{0.9}
	 \caption{Experimental results on CTW-1500.}\label{table:CTW1500}
	\begin{tabular}{m{2.5cm}<{\centering}
			|m{0.75cm}<{\centering}
			|m{0.6cm}<{\centering}
			m{0.6cm}<{\centering}
			m{0.6cm}<{\centering}
			m{0.5cm}<{\centering}}
		\toprule
		Methods& Ext& R& P & F & FPS\\
		\hline
		TextSnake \cite{TextSnake} &Syn&\textbf{85.3} &67.9 & 75.6 &- \\
		CSE \cite{CVPR19_CSE} &MLT&76.1 &78.7 & 77.4&0.38\\
		LOMO\cite{CVPR19_LOMO} &Syn&76.5 &85.7 & 80.8&4.4\\
		ATRR\cite{CVPR19_ATRR} &Sy-&80.2 &80.1 & 80.1 &-\\
		SegLink++ \cite{SegLink++} &Syn&79.8 &82.8 & 81.3 &-\\
		TextField \cite{TextField}&Syn&79.8 &83.0 & 81.4&6.0\\
		MSR\cite{MSR} &Syn&79.0 &84.1 & 81.5&4.3 \\
		PSENet-1s  \cite{CVPR19_PSENet} &MLT&79.7 &84.8 & 82.2&3.9\\
		DB \cite{DB}&Syn&80.2 &86.9 &83.4&{22.0}\\
		CRAFT \cite{CRAFT} &Syn&81.1 &86.0 &83.5&-\\
		TextDragon \cite{TextDragon} &MLT+&82.8 &84.5 &83.6&-\\
		PAN \cite{PSENet_v2} &Syn&81.2 &86.4 &83.7&\textbf{39.8}\\
		ContourNet~\cite{ContourNet}&-&84.1&83.7&83.9 &4.5\\
		DRRG~\cite{DRRG} &MLT&83.02&85.93&84.45&-\\
		TextPerception\cite{TextPerception}&Syn&81.9&87.5&84.6&-\\
		\hline
		\textbf{Ours}&-&80.57 &87.66 &83.97&12.08\\
		\textbf{Ours}&Syn &81.45 &\textbf{87.81}& 84.51&12.15\\
		\textbf{Ours}&MLT &83.60&86.45& \textbf{85.00}&12.21\\
		\bottomrule
	\end{tabular}
    \end{center}%
\vspace{-2.5em}
\end{table}

\noindent\textbf{Total-Text}. In testing, the threshold $ th_s $ is set to 0.825. The quantitative results are listed in Tab.~\ref{table:TotalText}. From Tab.~\ref{table:TotalText}, we can find that our method achieves $87.37\%$ in terms of F-measure when pre-trained on SynthText, and $87.85\%$ in the item of F-measure when pre-trained on MLT17. Obviously, our method significantly outperform other methods with a great margin. From the visible results in Fig.$ \, $\ref{fig:result_vis} (a) and Fig.$ \, $\ref{fig:result_vis} (b), we can observe that our method can precisely detect word-level irregular texts. 

\noindent\textbf{CTW-1500}.  In testing, the threshold $ th_s $ is set to 0.8. Representative visible results are shown in Fig.~\ref{fig:result_vis} (c) and (d), which indicate our method precisely detects boundaries of long curved text with line-level. The quantitative results are listed in Tab.$ \ $\ref{table:CTW1500}. Compared with the previous sate-of-the-art methods~\cite{DB,PSENet_v2,ContourNet}, our approach achieves promising performance of both precision ($87.81\%$) and F-measure ($ 85.0\% $). Specifically, our method greatly outperforms TextSnake~\cite{TextSnake}  and DB~\cite{DB} on CTW-1500 in item of F-measure by $9.4\%$ and $1.6\%$, respectively.

\begin{table}[htbp]
	\begin{center}
	\renewcommand{\arraystretch}{0.9}
	\caption{Experimental results on MSRA-TD500.}
	\label{table:TD500}
	\begin{tabular}{c|cccc}
		\toprule
		Methods& R& P & F&FPS\\
		\hline
		SegLink \cite{SegLink}  &70.0 &86.0 &77.0&8.9\\
		PixelLink \cite{PixelLink} &73.2 &83.0 & 77.8&-\\
		TextSnake \cite{TextSnake} &73.9 &83.2 &78.3&1.1\\
		TextField \cite{TextField}&75.9 &87.4 & 81.3&5.2\\
		MSR\cite{MSR} &76.7 &87.4 &81.7&-\\
		FTSN \cite{FTSN} &77.1 &87.6 &82.0&-\\
		LSE\cite{CVPR19_LSA} &81.7& 84.2 &82.9&-\\
		CRAFT \cite{CRAFT} &78.2 & 88.2 &82.9&8.6\\
		MCN \cite{MCN} &79 &88 &83&-\\
		ATRR\cite{CVPR19_ATRR} &82.1 &85.2 & 83.6&-\\
		PAN \cite{PSENet_v2}&83.8 &84.4  &84.1&30.2\\
		DB\cite{DB}&79.2&\textbf{91.5} &84.9&\textbf{32.0}\\
		DRRG~\cite{DRRG}&82.30&88.05&85.08&-\\
		\hline
		\textbf{Ours}~(Syn) &80.68 &85.40& 82.97&12.68\\
		\textbf{Ours}~(MLT) &\textbf{84.54} &86.62&\textbf{85.57}&12.31\\
		\bottomrule
	\end{tabular}
    \end{center}%
\vspace{-2.0em}
\end{table}

\noindent\textbf{MSRA-TD500}. In testing, the threshold $ th_s $ is set to 0.925. The quantitative comparisons with other methods on this dataset is listed in Tab.$ \, $\ref{table:TD500}. From Tab.~\ref{table:TD500}, we can conclude that our method
successfully detects long text lines of arbitrary orientations and sizes.  Notably, our method achieves $85.57\%$ in terms of F-measure, which outperforms other sate-of-the-art methods, such as DB~\cite{DB}, DRRG~\cite{DRRG}, \etc.


\section{Conclusion} \label{Conclusion}
In this paper, we propose a novel adaptive boundary proposal network for arbitrary shape text detection, which adopt an boundary proposal model to generate coarse boundary proposals, and then adopt an adaptive boundary deformation model combined with GCN and RNN to perform iterative boundary deformation to obtain the more accurate text instance shape. Extensive experiments show that the proposed method can precisely detects the boundaries of arbitrary shape text in challenging datasets. In future study, we are interested in  developing a real time method for text of arbitrary shapes based on current work.

\vspace{0.3em}
\noindent\textbf{Acknowledgements.} This work was supported in part by the National Key R\&D Program of China (2020AAA09701), National Natural Science Foundation of China (62006018, 62076024).

{\small
\bibliographystyle{ieee_fullname}
\bibliography{egbib}
}

\end{document}